\documentclass[conference]{IEEEtran}

\usepackage{graphicx,times,amsmath} 

\IEEEoverridecommandlockouts    

\begin{document}

\title{\ \\ \LARGE\bf Using Soft Constraints To Learn\\ Semantic Models Of Descriptions Of Shapes
\thanks{Sergio Guadarrama and David P. Pancho are with the European
Centre for Soft Computing, Mieres (Asturias), Spain (phone: +34 985456545;
email: \{sergio.guadarrama, david.perez\}@softcomputing.es).}}

\author{Sergio~Guadarrama, {\it Member, IEEE,} and  David P. Pancho}


\maketitle

\begin{abstract}

The contribution of this paper is to provide a semantic model (using soft
constraints) of the words used by web-users to describe objects in a language game; a game in which one user describes a selected object
of those composing the scene (see figure \ref{fig:complexDesc}), and another
user has to guess which object has been described. The given description needs
to be non ambiguous and accurate enough to allow other users to guess the
described shape correctly.

To build these semantic models the descriptions need to be analyzed to
extract the syntax and words' classes used (see
\cite{PanGua:2010:Syntax_learning} for details). We have modeled the meaning of
these descriptions using soft constraints as a way for grounding the meaning.

The descriptions generated by the system took into account the context of the
object to avoid ambiguous descriptions, and allowed users to guess the
described object correctly 72\% of the times.

\end{abstract}

\section{Introduction}


Language can be seen as a system learnt and used by humans for communicating
and learning, which covers a wide range of their daily activities. It is a
social phenomenon resulting in an evolving system of great complexity.
Language is inextricably linked to human capability to converse, learn, reason
and make decisions in an environment of imprecision, uncertainty and lack of
information. It is viewed here as a complex reality to be represented step by
step, in an incremental fashion.

In that respect, the most relevant feature of language is its ``meaning'' (its
``use'' according to Wittgenstein
\cite{Wittgenstein:1981:Philosophical_Investigations}), and that does not only
include the meaning/use of isolated words, but also the meaning/use of
expressions as a whole. In general, the meaning/use of the words integrating
an expression is only grasped in relation with the other words and within the
meaning/use of the expression as a whole in a context
\cite{Searle:1979:Expression_and_meaning:_studies_in_the_theory_of_speechacts}.

\begin{figure}[!htb]
\begin{center}
\centerline{\includegraphics[width=.7\linewidth]{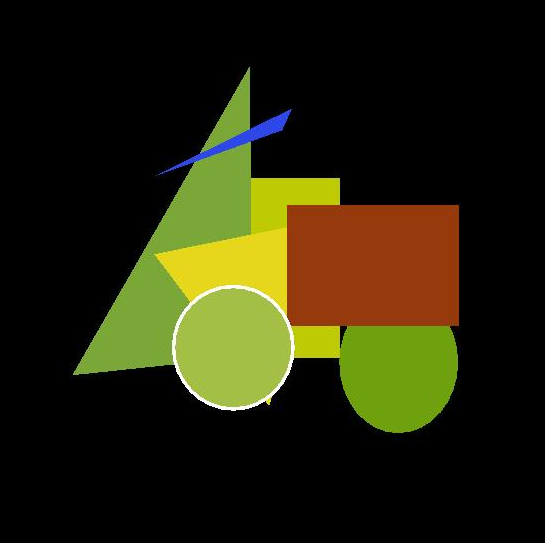}}
\end{center}
\caption{The green circle in the front} \label{fig:complexDesc}
\end{figure}

This work is part of  an ongoing project called Smart-Bees. Smart-Bees is a
project which aims to study how machines can learn and communicate in
human-like ways, from a Computing with Words, Actions and Perceptions (CW-AP)
perspective
\cite{Gua:2007:A_Contribution_to_Computing_with_Words_Perceptions,Mendel-etal:2010:What_computing_with_words_means_to_me}.
Users share a common environment and play different ``language-games'', in
this case they share a blackboard with geometric shapes of different colors,
sizes and positions, and play guessing and describing games. In the describing
game, given an image with one selected object that users try to describe the selected object to other
users in a non-ambiguous way  (see figure \ref{fig:complexDesc}). In the guessing game, on the contrary, given a
description and an image users try to guess which object was described.

This project has its roots in Wittgentstein's ideas  about Meaning and
Language
\cite{Wittgenstein:1981:Philosophical_Investigations, Wittgenstein:1958:Blue_and_Brown_books},
Zadeh's ideas on Linguistic Variables, Computing with Words and Generalized
Constraints
\cite{Zadeh:1975:The_concept_of_Linguistic_Variable,Zadeh:1999:From_Computing_with_numbers_to_Computing_with_Words,Zadeh:1996:Test-score_semantics_for_natural_languages_and_meaning_representation_via_PRUF,zadeh:2006:A_new_frontier_in_computation,zadeh:2005:Toward_a_generalized_theory_of_uncertainty_GTU_an_outline},
Trillas' ideas on Words and Fuzzy Sets
\cite{Tri:2006:On_the_use_of_words_and_fuzzy_sets, TriGua:2005:What_about_fuzzy_logics_linguistic_soundness},
Roy's ideas on meaning grounding
\cite{RoyPentland:2002:Learning_words_from_sights_and_sounds,Roy:2002:Learning_visually_grounded_words,GorniakRoy:2004:Grounded_semantic_composition_for_visual_scenes,Roy:2005:Grounding_words_in_perception_and_action,OrkinRoy:2007:The_restaurant_game},
and Guadarrama's works on Computing with Words, Actions and Perceptions
\cite{Gua:2007:A_Contribution_to_Computing_with_Words_Perceptions,GuaGar:2008:Concept_Analyzer_A_tool_for_analyzing_fuzzy_concepts,Guadarrama:2009:Computing_with_actions}.


To learn semantic models from descriptions of shapes given by users several
steps are needed; to collect descriptions of shapes from web-users; to learn
the lexicon and syntax used in that descriptions; to link that lexicon and
syntax with the features of the shapes to learn the semantics; to generate new
descriptions using the syntax and semantic learned; to test them with
web-users. The final goal is to learn concepts, words and some sort of syntax
and semantics, building a model grounded in the shared perceptions.

The contribution of this paper is to provide a semantic model (using soft
constraints) of the words used by web-users to describe objects in a language game; a game in which one user describes the selected
object among those composing the scene, and another user has to guess which
object has been described (see figure \ref{fig:complexDesc}). The given description needs to be non ambiguous and accurate enough to allow other users to
guess the described object correctly.


So far the system has 40 registered users, from 15 different countries, who
had provided 360 descriptions using 150 different words and had allowed the
system to learn some lexicon (30 words), some syntax (20 patterns), and some
semantics (7 word's classes grounded) for the shape description task. The
method proposed in this paper is performing quite well obtaining 100\% correct
spelled words, 88\% syntactically correct sentences and 72\% of semantically
correct sentences; however users spelled on average correctly 97\% of the
words, wrote 93\% of syntactically correct sentences and provided 75\% of
semantically correct sentences.


The rest of the paper is structured as follows. In Section \ref{sec:related}
we describe related works and compare them with this one. In Section
\ref{sec:semantic} we present our model of the meaning of words based on soft
constraints, in Section \ref{sec:learning} we present the learning algorithm
to learn the soft constraints from the data, and in Section
\ref{sec:generating} we present how the new descriptions are generated.
Finally Section \ref{sec:results} presents the main results and Section
\ref{sec:conclusions} the main conclusions of the paper.

\section{Related works}
\label{sec:related}

The experiment presented here has been inspired in the DESCRIBER system done
by Roy in \cite{Roy:2002:Learning_visually_grounded_words}, where he presented
a similar problem of learning the descriptions provided by one user about an
scene composed by non-overlapping squares and rectangles. Nevertheless we have
turned the experiment more realistic in several aspects: allowing different users to provide
descriptions (web-users not familiar with the experiment), including more kind
of shapes (triangles, circles, ovals) and allowing them to overlap (making
harder the segmentation and the descriptions). Also it is important to remark
that in the previous experiment the only user was a native English speaker,
who provided very consistent descriptions, without spelling or syntactical errors and very few ambiguous descriptions, while in own experiment we have a variety of users from 15 different countries with only few being non-English native speakers.

In that case the system they proposed was using bi-grams for the syntax
learning and Gaussian mixtures for the semantical learning, obtaining a 81.3\%
of correct descriptions. But they have only one user who spelled correctly
100\% of the words, wrote 100\% of syntactically correct sentences and was
able to provide 89\% of semantically correct sentences, in comparison with our
case that we start with 40 users from 15 different countries who on average
spelled correctly 97\% of the words, wrote 93\% of syntactically correct
sentences, and were able to provide only 75\% of semantically correct
sentences.

The problem of learning grounded words has been studied in
\cite{Roy:2002:Learning_visually_grounded_words,GorniakRoy:2004:Grounded_semantic_composition_for_visual_scenes,Roy:2005:Grounding_words_in_perception_and_action,Arensetal:2008:Conceptual_representations,LiangEtal:2009:Learning_semantic_correspondences}, the problem of social learning have been studied in
\cite{Steels:2001:Language_games_for_autonomous_robots,SteelsKaplan:2002:Aibo_first_words}.
The need to extend Fuzzy Logic to cope with the problems of CW-AP has been
recently remarked in
\cite{Gua:2007:A_Contribution_to_Computing_with_Words_Perceptions,Zadeh:2008:Is_there_a_need_for_fuzzy_logic,
Zadeh:2009:Computing_with_words_and_perceptions-a_paradigm_shift,
PraTriGuaRen:2007:On_fuzzy_set_theories,
Tri:2009:On_a_model_for_the_meaning_of_predicates}.

These previous works may be contrasted with this one in two aspects. First, we
see language learning as an integrated process where sensory-action learning,
social learning and supervised learning are interleaved and combined. Second,
we start from a multi-user perspective, where different users and agents
interact and share their knowledge.

The main difference of our approach with respect to other published works is
the path taken, that is, the movement from manipulation of measurements to
manipulation of perceptions, and  from syntax-based systems to semantics-based
systems. Other approaches underestimate the importance of imprecision inherent
in language \cite{TriRenGua:2006:Fuzzy_Sets_vs_Language} and in perception
\cite{Zadeh:2001:A_new_direction_in_AI} and have tried to reduce
it to simple forms of uncertainty, instead of dealing with it through a
general theory of uncertainty
\cite{zadeh:2005:Toward_a_generalized_theory_of_uncertainty_GTU_an_outline}.

\section{Semantic models of words using soft constraints}
\label{sec:semantic}

The meaning of a word is its use in language, and therefore it is
context-dependent. Actually, words are grounded in actions and perceptions,
and its use is learnt in a semi-supervised environment. Let us
summarize the main problems that have been faced in this work:
\begin{itemize}
 \item There are several misspelled words, some rare-words, that need to be corrected or discarded.
 \item Different users use different words, different syntactic patterns and in some cases with different meanings.
 \item The same object can be described in many different ways, depending on
 the context and on the intention of the user.
 \item Descriptions made should be understandable by other
 users, so it should be truthful, precise, context-relevant and non-ambiguous.
 \item There are few examples for each word, and not all the possible combinations are seen.
 \item Semi-supervised problem, only a small part of the data can be  supervised.
\end{itemize}

To collect human descriptions of shapes and to test the results, we have set
up an interactive website. The system learn from the descriptions provided by
humans and use the method described in this paper to produce its own
descriptions.
\begin{center}\emph{https://www3.softcomputing.es/smart-bees}\end{center}

These are some examples of simple descriptions given by users: ``the green
rectangle'',  ``big green triangle'', ``brown rectangle''. And these are
examples of compound descriptions: ``light green rectangle at the bottom'',
``pink circle behind dark green square'', ``the shape under the green one'',
``light blue circle in the middle'', ``orange circle behind the yellow
circle'', ``green small square in the background'', ``the dark orange
rectangle behind the triangle''.

In this paper we will focus on the semantic learning, once the lexicon and
syntax have been learnt (see paper \cite{PanGua:2010:Syntax_learning} also
presented in this conference), and on the generation of new descriptions and
their validation. Using the results from \cite{PanGua:2010:Syntax_learning} we
transformed the original problem of pairs of descriptions and shapes into a
problem of sets of pairs of words and shapes. In which each set represents a
class of words extracted by the syntax.

Thus, each shape will be associated with all the words used in the
description, and therefore it will have multiple labels attached. Given a set
of words' classes associated with a set of objects, we need to learn when each
word of the class is used based on the features of the shapes (see section
\ref{sec:learning}). This problem is similar to a multiple labeling
problem, in which for each object and each words' class we need to decide
which labels are applicable and to which degree.

Given a set of pairs of words (taken as labels) and shapes (taken as objects)
we need to learn why, when and how each label is used according to the
features of the shapes, to the relations between shapes and to the grammatical
rules. To calculate the degree of matching between a description and a selected
object in a scene is very important, since it will be used later to calculate
the degree of ambiguity, by comparing it with the matching degrees between the
description and the other shapes forming the scene.

\subsection{Modeling the Meaning of Propositions}
\label{sec:modelling_propositions}

As Zadeh suggested in
\cite{Zadeh:2009:Computing_with_words_and_perceptions-a_paradigm_shift} and in
\cite{Zadeh:2004:Precisiated_natural_language}
 every
proposition can be represented by a generalized constraint.
\[\mbox{``p''} \Rightarrow \mbox{X is R}.\]
Where $X$ is a relevant variable constrained by $R$

{Example}
\[\mbox{``John is Tall''} \Rightarrow Height(John) \mbox{ is } Tall\]

Where $Height(John)$ is a projection of some attributes of John. And $Tall$ is
a constraint on the values of the attributes of John.

\subsection{Modeling the Meaning of Descriptions}
\label{sec:modelling_descriptions} In our case, given a description of an
object $'x'$ it can be represented by a set of constraints. For example:
\begin{itemize}
\item ``The blue square''
\[Color(x) \mbox{ is } Blue \mbox{ and } Shape(x) \mbox{ is } Square\]
\item ``The big dark green triangle in the background''
\[\begin{array}{c} Color(x) \mbox{ is } Dark\;Green \mbox{ and } Shape(x) \mbox{ is } Triangle\\
  \mbox{ and } Position(x) \mbox{ is } Background
  \end{array}\]
\end{itemize}

Where $Color$, $Shape$ and $Position$ are projections of the features of $x$,
and $Blue$, $Square$, $Dark\; Green$, $Triangle$, $Background$  are
constraints on the values of the projected features.

Thus from the descriptions provided by the users and their corresponding
images, the system learns which projections are associated with which words,
and which constraints represent their meaning.

\subsection{Learning process}
\label{sec:learning_process}

The general phases of the learning process are listed bellow:
 \begin{itemize}
   \item Learning the Lexicon: in this phase it is needed to select relevant words and filter misspelled words (it is presented in \cite{PanGua:2010:Syntax_learning})
   \item Learning the Syntax: in this phase it is required to group words according to their role in
   the sentence, and learn a grammar (it is presented in \cite{PanGua:2010:Syntax_learning}, and briefly shown in \ref{sec:wordsclasses}).
   \item Segmentation of images to extract objects and features, and pair the segmented objects with descriptions (this is presented in section \ref{sec:segmentation}).
   \item Learning the Semantics: Generate a model for each word belonging to the cluster in the
   projected space according to the features selected (this phase is presented in section \ref{sec:learning}.
   \item Generation sentences: in this phase syntactically and semantically correct sentences are generated for new images (this phase is presented in section \ref{sec:generating}).
   \item Evaluation of results: Once all the sentences are generated an evaluation process is performed, in which the users try to understand the sentences and select the corresponding object (this phase is presented in section \ref{sec:results}).
 \end{itemize}

Let us recall the results of the lexical and syntax learning phases from the
paper \cite{PanGua:2010:Syntax_learning}. Words with frequency smaller that 10
have been filtered, and remained 30 words which after clustering formed 7
words' classes (shown in \ref{sec:wordsclasses}), and generated a syntax
composed by 20 patterns shown in table \ref{tab:freqpatterns}.

\subsection{Classes of words} \label{sec:wordsclasses}
{\small
\noindent Class 1 = \{ THE, A  \}\\
Class 2 = \{ BACKGROUND, FRONT \}\\
Class 3 = \{ CIRCLE, OVAL, TRIANGLE, RECTANGLE, ELLIPSE, SQUARE \} \\
Class 4 = \{ ON, IN, AT, BEHIND \}\\
Class 5 = \{ LIGHT, BIG, DARK \}\\
Class 6 = \{ TOP, BOTTOM, RIGHT, LEFT \}\\
Class 7 = \{ PINK, BLUE, GREEN, ORANGE, RED, YELLOW, PURPLE, VIOLET, BROWN \} }

\begin{table}[!htb]
\centering \caption{Most frequent patterns}\label{tab:freqpatterns}
\begin{tabular}{c|l}
  \hline\hline
  Frequency & Pattern \\
  \hline
    18.89\%     &     7     3 \\
    6.94\%     &     1     7     3\\
    6.39\%     &     1     3\\
    5.83\%     &     3 \\
    3.89\%     &     7     3     4     1     2 \\
    3.33\%     &     5     7     3 \\
    3.06\%     &     2     7     3 \\
    2.50\%      &     1     7     3     4     1     6\\
    2.22\%     &     7     3     4     1     6 \\
    1.66\%     &     7 \\
    1.11\%     &     6     3 \\
    1.11\%    &     1     7     3     1     7     3\\
    0.83\%   &     1     7 \\
    0.83\% &     1     5     7     3\\
    0.83\% &     2     5     7     3\\
    0.83\% &     3     4     1     2\\
    0.83\% &     3     4     1     6\\
    0.83\% &     7     3     4     1     3\\
    0.83\% &     1     3     4     1     6     6\\
    0.83\% &     5     7     3     4     1     2\\
  \hline\hline
\end{tabular}
\end{table}

\section{Learning the Semantics}
\label{sec:learning}

The system needs to learn why those specific words were used to describe that object in that context (image). For that we analyzed images to
segment and extract objects and measure their features. We used a scaffolding
learning: starting from simple descriptions before learning compound
descriptions; of the 360 descriptions with all their words in the lexicon,
75\% are simple and 25\% are compound.

Words belonging to the same class have different meanings; for example given a
class of words = \{'BLUE', 'RED', 'GREEN', 'YELLOW',..\} we assume that each
word have a different meaning, and therefore should be represented by
different model, even though, in some cases different words can be applied to
the same object to some extent.

\subsection{Shapes' segmentation} \label{sec:segmentation}
A fuzzy edge detector was used to find the edges of the shapes, then using a filling transformation to
found the regions inside the edges, and finally using a color-based clustering
and an overlapping detection we grouped the regions into shapes. After
obtaining a set of candidate shapes -- comprised by a set of pixels -- they
were matched with the selected object and its corresponding description.

For each shape a set of 20 features were measured, including:
\begin{itemize}
 \item Average RGB: Red, Green, Blue.
 \item Average YCbCr: Y is the luma component, and Cb and Cr are the blue-difference and red-difference chroma components.
 \item Bounding Box: Coordinates of the bounding box.
 \item Height and width.
 \item Center of gravity: position of the center of gravity.
 \item Bounding Ellipse: Orientation and size of the bounding ellipse.
 \item Major Minor: length of the major and minor axis of the bounding
 ellipse.
 \item Extension: proportion of the bounding box filled.
 \item Height to width ratio.
 \item Area: number of pixels.
 \item Holes: proportion of holes in the object.
 \item ...
\end{itemize}


\subsection{Multi-classification problem}

It is important to notice that different users describe differently the same
objects, even they used different words and different syntax. So the training
data could contain different labels for the same object or not label at all.
Some objects have only labels for some of the word's classes but nor for all;
for example ``The blue square'' only specify that the color is blue and the
shape is square but say nothing about the size or position of the object
described.

There are also many objects that have not being described by any user, so we
also have many un-labeled objects.

\begin{figure}[!htb]
\begin{center}
\includegraphics[width=.7\linewidth]{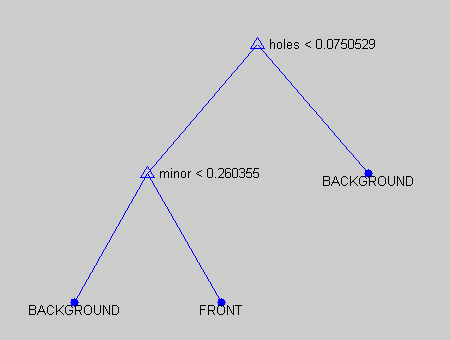}
\caption{Fuzzy Decision Tree for Class 2 (Depth)} \label{fig:class-depth}
\end{center}
\end{figure}

The system learn which projection (relevant features) is appropriate for each
class of words and which constraints (relevant values) are associated with
each word. For every class of words we assume that one projection is shared by
all the words in the class. For every word in a class we assume that it is
represented by one constraint over the projection of the class.

To obtain a robust classifier in despite of the aforementioned problems we have decided to use fuzzy decision trees for their robustness and flexibility. And also because they also do feature selection during the learning process
\subsection{Fuzzy decision Trees}
\label{sec:decisiontrees}

A different set of features could be relevant for each class of words. So we
used  fuzzy decision trees \cite{Janikow:1998:Fuzzy_decision_trees} to classify the objects according to their labels and cross validation to prune the tree and select the most relevant features. In figure \ref{fig:class-depth} can be seen the
fuzzy decision tree of Class 2.

The features selected for each class are the following:

\begin{center}
\begin{tabular}{c|l}
 \hline\hline
 Class & Features \\\hline
 Class 1 & --\\
 Class 2 & Holes Minor \\
 Class 3 & Ext HW-ratio\\
 Class 4 & --\\
 Class 5 & G \\
 Class 6 & X Area  \\
 Class 7 & Cr Cb\\\hline\hline
\end{tabular}
\end{center}

From the features selected we can see that none is related to Class 1 nor to Class 3, that means that from the current features their meaning remains
unground (or unlearned). This is due to the fact that those classes are more related to the syntax that to the semantics, nevertheless the fuzzy decision tree learns that the most frequent word should be used by default.

\begin{figure}[!htb]
\begin{center}
\includegraphics[width=.7\linewidth]{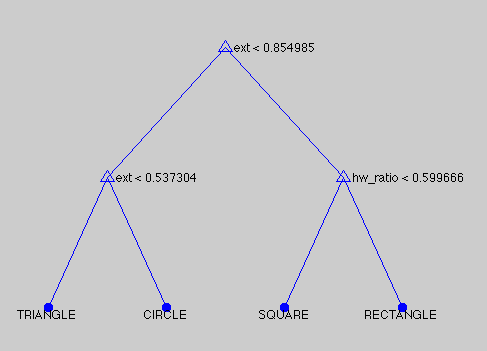}
\caption{Fuzzy Decision Tree for Class 3 (Shapes)} \label{fig:class-shapes}
\end{center}
\end{figure}

\begin{figure}[!htb]
\begin{center}
\includegraphics[width=.65\linewidth]{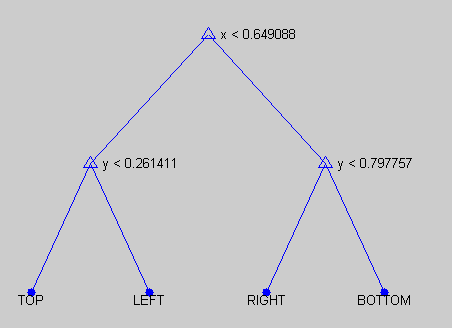}
\caption{Fuzzy Decision Tree for Class 6 (Positions)}
\label{fig:class-positions}
\end{center}
\end{figure}

\begin{figure}[!htb]
\begin{center}
\includegraphics[width=.7\linewidth]{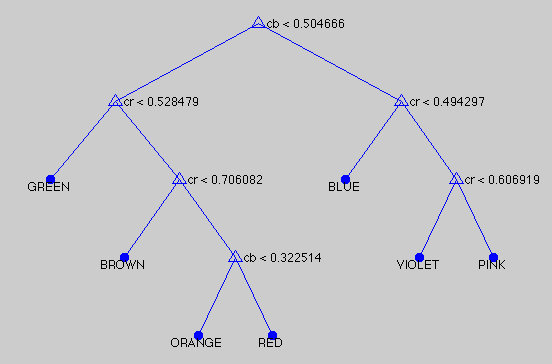}
\caption{Fuzzy Decision Tree for Class 7 (Colors)} \label{fig:class-colors}
\end{center}
\end{figure}

The decision trees for each class are the following:

\begin{itemize}
 \item Class 1: If true then 'THE'
 \item Class 2: See figure \ref{fig:class-depth}
 \item Class 3: See figure \ref{fig:class-shapes}
 \item Class 4: If true then 'IN'
 \item Class 5: If $g\leq 0.64$ then 'LIGHT' else 'DARK'
 \item Class 6: See figure \ref{fig:class-positions}
 \item Class 7: See figure \ref{fig:class-colors}
\end{itemize}

\subsection{Fuzzy Labels}
\label{sec:fuzzylabels}

\begin{figure}[!htb]
\begin{center}
\includegraphics[width=.8\linewidth]{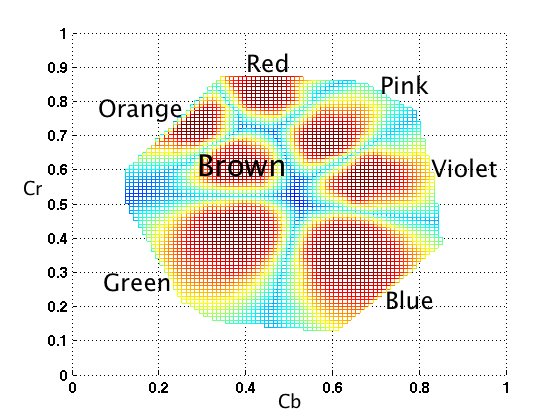}
\caption{Fuzzy Labels for Class 7 (Colors)} \label{fig:fuzzy-clusters-colors}
\end{center}
\end{figure}

\begin{figure}[!htb]
\begin{center}
\includegraphics[width=.8\linewidth]{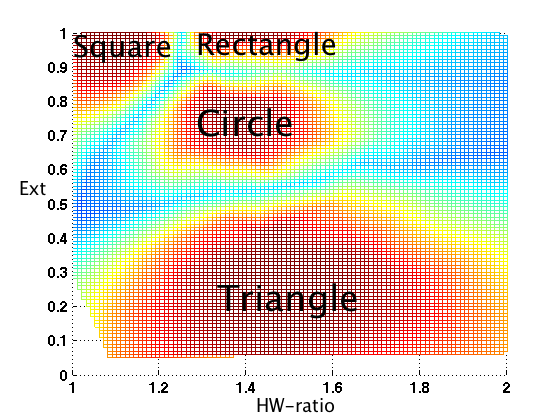}
\caption{Fuzzy Labels for Class 3 (Shapes)} \label{fig:fuzzy-clusters-shapes}
\end{center}
\end{figure}

Once the fuzzy decision trees are built for each word's class we can calculate
the degree of matching between every object and every word obtaining a soft
constraint for each label. For example in the case of class 7 (colors) and
class 3 (shapes) we obtain the fuzzy labels plotted in figures
\ref{fig:fuzzy-clusters-colors} and \ref{fig:fuzzy-clusters-shapes}
respectively.

\subsection{Degree of matching of descriptions}
\label{sec:matchingdegree}

The degree of matching between one description and one object depends on the degree of matching of each word composing the description and on an aggregation function (in our case the minimum).

Once the projections (relevant features) and the soft constraints (fuzzy labels) have been learnt for each word's class we can transform every description into generalized constraints using the syntax, as follows:
\begin{table}[!htb]
\centering
\begin{tabular}{c|c|c}
\hline\hline The & blue & square \\\hline
 class1 & class7 & class3\\\hline
  -- & $CrCb(x)$ is $Blue$ & $ExtHW_{ratio}(x)$ is $Square$\\\hline
  $\mu_{The}$ & $\mu_{Blue}$ & $\mu_{Square}$\\
  \hline\hline
 \end{tabular}
\end{table}

From that we can calculate the degree of matching $\mu_{M}$ between an
object $x$ and a description $D$ by:
\[\mu_{M}(x,D)= \bigcap\mu_{label_i}(x) \;; \forall\; label_i \in pattern(D)\]
where $pattern(D)$ is the sequence of labels of a given description, and
$\mu_{label_i}$ is the fuzzy label representing each word of the description.

\subsection{Degree of ambiguity}
The degree of ambiguity of one description in one scene depends on the degrees of matching between the description and the objects of the scene. Because if there are more than one object with high degree of matching then the description could refer to various objects and be ambiguous.

In every scene there are several objects, and any given description can be
ambiguous if it is applicable to several of these objects. We can calculate
the degree of ambiguity $\sigma_{A}$ of a description $D$ in an scene $S$ by:

\[\sigma_{A}(D,x,S) = \underset{y\neq x, y \in S}{Sup}\mu_{M}(y,D))\]

where $x$ is the object with highest degree of matching and  $\mu_{M}(y,D)$ represents the degree of matching between the description $D$ and the other objects $y\neq x$ present in the scene. Thus the higher the degree of matching with the other objects the higher the degree of ambiguity, because the description would not be discriminative enough.

\section{Generating descriptions}
\label{sec:generating}

For generating descriptions the system will look for short, truthful and
non-ambigous descriptions, and will follow the next algorithm:
\begin{enumerate}
 \item Given an scene with one selected object.
 \item Segment it, extract the objects and their features.
 \item Get the most frequent short syntax pattern. \label{eun:most}
 \item For each word's class find the label with the highest degree of matching.
 \item Build the description and calculate the degree of ambiguity.
 \item If the description is non-ambiguous return the description with the highest degree of matching; else go to step \ref{eun:most}) and look for the next pattern and repeat the process.
\end{enumerate}

\begin{figure}[!htb]
\begin{center}
\centerline{\includegraphics[width=.7\linewidth]{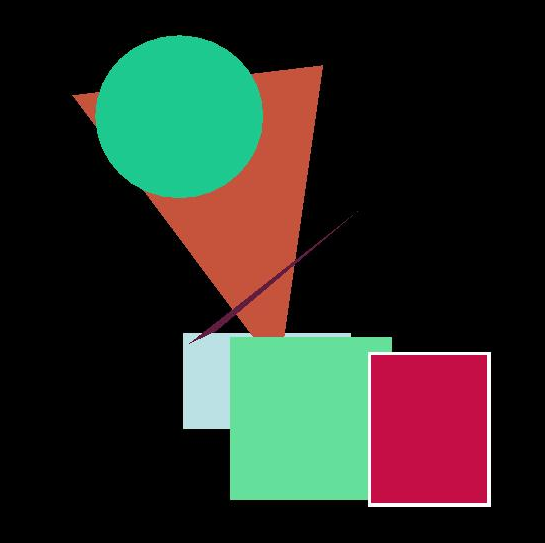}}
\caption{The red rectangle} \label{fig:simpleDesc}
\end{center}
\end{figure}

For example, in the scene seen in figure \ref{fig:simpleDesc}, the system
segment it and found 7 objects with their 20 features, starting by most
frequent short pattern (1 3 7) it calculates the degree of matching for each
label in Class 1, Class 3 and in Class 7;  it finds that $The \in Class 1$,
$Rectangle \in Class 3$ and $Red \in Class 7$ have the highest degree of
matching
\[\begin{array}{rl}
D=& \mbox{`The red rectangle'}\\
\mu_{M}(x,D)= & \min(\mu_{The}(x),\mu_{Red}(x),\mu_{Rectangle}(x))\\
=& \min(1,0.68,0.74) =0.68\\\\
\sigma_{A}(D,x,S) =& \underset{y\neq x, y \in S}{Sup}\mu_{M}(y,D)) =  0.11
\end{array}\]

Nevertheless, in the scene seen in figure \ref{fig:complexDesc}, when the
system calculates the degree of matching starting by most frequent short
pattern (1 3 7) it finds out that the degree of ambiguity is high.
\[\begin{array}{rl}
D=& \mbox{`The green circle'}\\
\mu_{M}(x,D)= & \min(\mu_{The}(x),\mu_{Green}(x),\mu_{Circle}(x))\\
=& \min(1,0.78,0.57) =0.57\\\\
\sigma_{A}(D,x,S) =& \underset{y\neq x, y \in S}{Sup}\mu_{M}(y,D)) =  0.53
\end{array}\]

But it turns out that the ambiguity degree is also high, thus the system keep
trying with other patterns until it finds one with lower degree of ambiguity
(1 3 7 4 1 2) while maintaining a high degree of matching, in this case:

\[\begin{array}{rl}
D=& \mbox{`The green circle in the front'}\\
\sigma_{M}(x,D)= & \min(\mu_{The}(x),\mu_{Green}(x),\mu_{Circle}(x),\\
& \mu_{In}(x),\mu_{The},\mu_{Front}(x))\\
=& \min(1,0.78,0.57,1,1,0.61) =0.57\\\\
\sigma_{A}(D,x,S) =& \underset{y\neq x, y \in S}{Sup}\mu_{M}(y,D)) =  0.07
\end{array}\]

\section{Results}
\label{sec:results}

 To compare this work with the previous one
\cite{Roy:2002:Learning_visually_grounded_words} and to check the influence of
the different options considered in the paper we have defined three methods:
\begin{itemize}
\item Method 1: In this case we used the algorithm and features proposed in this paper but without using the degree of ambiguity to avoid ambiguous descriptions.
\item Method 2: In this case we used the algorithm and features proposed by Roy in his paper \cite{Roy:2002:Learning_visually_grounded_words}.
\item Method 3: In this case we used the algorithm and features proposed in this paper and used the degree of ambiguity to avoid ambiguous descriptions.
\end{itemize}

For the scene shown in figure \ref{fig:complexDesc} the descriptions generated
by the three methods are:

\begin{itemize}
\item Method 1: GREEN CIRCLE
\item Method 2: THE LIGHT GREEN CIRCLE
\item Method 3: THE GREEN CIRCLE IN THE FRONT
\end{itemize}

and for the scene shown in figure \ref{fig:simpleDesc} are:

\begin{itemize}
\item Method 1: THE RED RECTANGLE
\item Method 2: THE PINK RECTANGLE
\item Method 3: THE RED RECTANGLE
\end{itemize}

\begin{figure}[!htb]
\begin{center}
\includegraphics[width=1.1\linewidth]{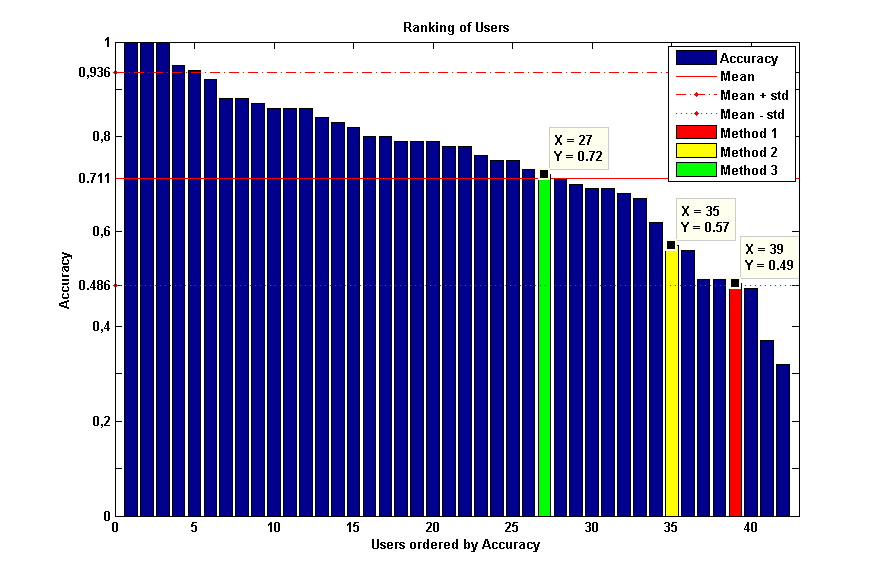}
\caption{Ranking users according to their performance}
\label{fig:ranking-user}
\end{center}
\end{figure}

After generating the descriptions for the 350 scenes using the three methods we included them in the web-page, so the users can try to guess which objects are being described. To warranty the fairness of the experiment the users don't know which descriptions are generated automatically by the system and which ones come from other users. Actually which description is shown to each user is selected randomly among all. Counting as correct that descriptions that other users guessed right we obtained the results showed in figure \ref{fig:ranking-user}.

In figure \ref{fig:ranking-user} can be seen that the Method 1 is
performing bellow average, it obtains 49\% of the descriptions correct, and it
is ranked \#39 which means that other 4 users are performing even worse. The
Method 2 is performing a little bit better obtaining 57\% of the descriptions
correct (while below the results presented in the previous work 81.3\%) and it
is ranked \#35. The Method 3 is performing quite well obtaining 72\% of the
descriptions correct (just a little bit over the average of users) and it is
ranked \#27.

\section{Conclusions}
\label{sec:conclusions}

So far the system has 40 registered users, from 15 different countries, who
had provided 360 descriptions using 150 different words and had allowed the
system to learn some lexicon (30 words), some syntax (20 patterns), and some
semantics (7 word's classes grounded) for the shape description task. The best
method it is performing quite well obtaining 100\% correct spelled words, 88\%
syntactically correct sentences and 72\% of semantically correct sentences;
despite the variety of users, who spelled correctly 97\% of the words, wrote
93\% of syntactically correct sentences and provided 75\% of semantically
correct sentences.

We have provided a semantic model (using soft constraints) of the words used
by web-users to describe objects for other users in a describing game. The
descriptions generated took into account the context of the object to avoid
ambiguous descriptions, allowing users to guess the described object
correctly. A future work is to study the construction of complex phrases,
those referring to more than one object.

With the approach taken in this work the possibility to study semantic models
for specific words used by specific users in specific contexts is opened. This
can be seen as a step in the development of Computing with Words whose
relevance have been highlighted by Zadeh in
\cite{zadeh:2006:A_new_frontier_in_computation,Zadeh:2009:Computing_with_words_and_perceptions-a_paradigm_shift}

\section*{Acknowledgment}

This work has been supported by the Foundation for the Advancement of Soft
Computing (ECSC) (Asturias, Spain), the Spanish Department of Science and
Innovation (MICINN) under program Juan de la Cierva JCI-2008-3531, and the
European Social Fund.

\bibliographystyle{IEEEtran}


\bibliography{E:/pubs/bib/xbib_paper,E:/pubs/bib/xbib_proc,E:/pubs/bib/xbib_book,E:/pubs/bib/xbib_otros}

\end{document}